%% file: main.tex
\documentclass[10pt,twocolumn,letterpaper]{article}
\usepackage[pagenumbers]{cvpr}      
\usepackage[accsupp]{axessibility} 
\usepackage{tabularx}
\usepackage{multirow}
\usepackage{makecell}
\usepackage[marginal]{footmisc}

\input{preamble}

\definecolor{cvprblue}{rgb}{0.21,0.49,0.74}
\usepackage[pagebackref,breaklinks,colorlinks,citecolor=cvprblue]{hyperref}

\def\paperID{9383} 
\def\confName{CVPR}
\def\confYear{2024}
\title{Efficient Multi-scale Network with Learnable Discrete Wavelet Transform \\ for Blind Motion Deblurring\vspace{-5.5mm}}
\author{Xin Gao$^{*}$$^{1,2}$\quad Tianheng Qiu$^{*}$$^{2,3,4}$\\
Xinyu Zhang$^{\dag}$$^{2}$ \;
Hanlin Bai$^{1}$ \;
Kang Liu$^{1}$ \; 
Xuan Huang$^{\dag}$$^{4}$ \; 
Hu Wei$^{4}$ \; 
Guoying Zhang$^{\dag}$$^{1}$ \; 
Huaping Liu$^{2}$ \vspace{-0mm}\\
\small $^{*}$Equal Contribution \qquad $^{\dag}$ Coresponding Author \vspace{1.8mm}\\
$^{1}$China University of Mining \& Technology-Beijing \quad $^{2}$Tsinghua University \\
$^{3}$University of Science and Technology of China\\
$^{4}$Hefei Institutes of Physical Science,Chinese Academy of Sciences
\vspace{-4mm}
\\
}
\begin{document}
\maketitle
\begin{abstract}
Coarse-to-fine schemes are widely used in traditional single-image motion deblur; however, in the context of deep learning, existing multi-scale algorithms not only require the use of complex modules for feature fusion of low-scale RGB images and deep semantics, but also manually generate low-resolution pairs of images that do not have sufficient confidence. In this work, we propose a multi-scale network based on single-input and multiple-outputs(SIMO) for motion deblurring. This simplifies the complexity of algorithms based on a coarse-to-fine scheme. To alleviate restoration defects impacting detail information brought about by using a multi-scale architecture, we combine the characteristics of real-world blurring trajectories with a learnable wavelet transform module to focus on the directional continuity and frequency features of the step-by-step transitions between blurred images to sharp images. In conclusion, we propose a multi-scale network with a learnable discrete wavelet transform (MLWNet), which exhibits state-of-the-art performance on multiple real-world deblurred datasets, in terms of both subjective and objective quality as well as computational efficiency.
Our code is available on \url{https://github.com/thqiu0419/MLWNet}.

\end{abstract}
\section{Introduction}
\begin{figure}[htbp]
\setlength{\abovecaptionskip}{-0.3cm}
\setlength{\belowcaptionskip}{-0.6cm}
    \begin{center}
    \includegraphics[width=1.\linewidth]{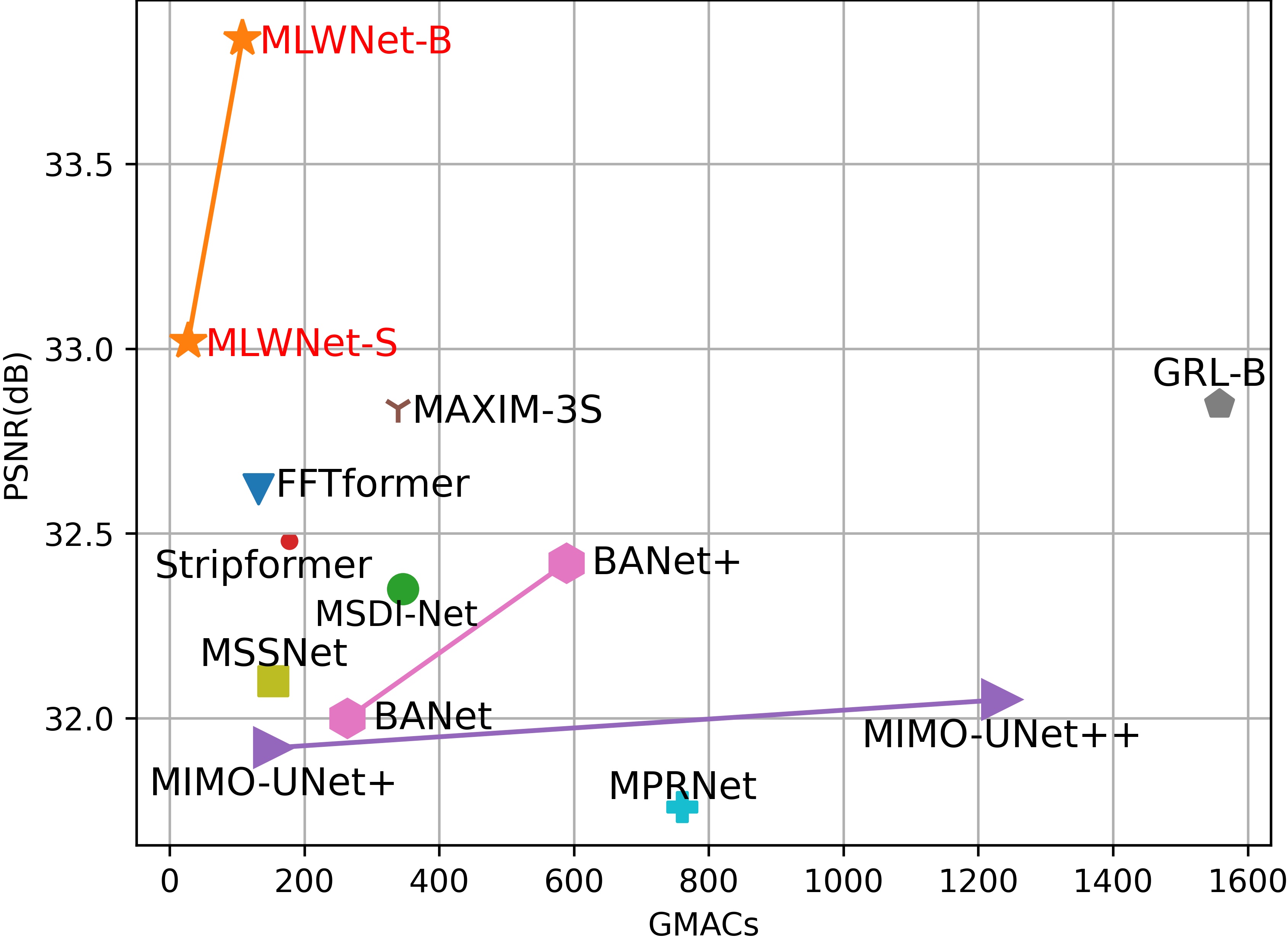}
    \end{center}
    \caption{Performance comparison on the RealBlur-J~\cite{rim2020real} test dataset in terms of PSNR and GMACs. 
    Our proposed MLWNet achieves superiority in comparison with other state-of-the-arts.}
    \label{fig:metrics}
\end{figure}

Most current top-performing single-image blind deblurring algorithms are based on DNNs, which can be structurally categorized into single-scale~\cite{kong2023efficient, chen2022simple, zou2021sdwnet, kupyn2019deblurgan, tsai2022banet, tsai2022stripformer} and multi-scale approaches~\cite{tao2018scale,zamir2021multi,cho2021rethinking,kim2022mssnet,li2022learning}. 
Compared to single-scale methods, multi-scale methods are built on the idea of moving from coarse-to-fine, and thus they decompose the challenging single-image blind deblurring problem into easier-to-solve sub-problems to restore the blurred image step-by-step.

Earlier DNNs exploiting the coarse-to-fine concept usually employ estimators of different scales to produce restored images gradually~\cite{tao2018scale,nah2017deep,park2020multi}. However, not only is the extracted semantic information not sufficiently representative, but the use of the same complexity at different scales may lead to redundant computations; More advanced multi-scale algorithms~\cite{cho2021rethinking} use a global encoder-decoder, which significantly reduces the algorithm's runtime, multiple upsampling and downsampling leads to insufficient restoration of detailed information, and therefore such algorithms require the introduction of additional fusion modules to encode input images for fusion with deeper semantics. In addition, existing multi-scale algorithms use multiple-input and multiple-output (MIMO) architectures, which require the introduction of manually constructed downsampled image pairs, and usually employ simple interpolation algorithms to generate low-resolution images for the sake of efficiency, which is obviously not reliable.

We note that the gradual insertion of images used by existing coarse-to-fine algorithms is redundant. Referring to numerous algorithms~\cite{lin2017feature,tan2020efficientdet,liu2018path,ghiasi2019fpn,kim2022mstr,wang2023yolov7} that are widely used in downstream tasks, the network starts with a single input and also extracts features at different scales. These features can be aggregated by simple feature fusion to ultimately obtain competitive results. That is, a DNN itself possesses the ability to learn effective features at different scales. This inspired us to design a multi-scale architecture that retains an original image as input and sequentially restores images at each scale in output stage. This is not only more theoretically sound, but also eliminates the time required to generate pairs of images of different resolutions, as well as the huge complexity increase brought about by fusion modules between RGB domain and deep features.

The coarse-to-fine algorithm has another inherent defect. During progressive restoration, the solution of the upper-level problem takes the solution of the lower-level as initialization~\cite{kim2022mssnet}. Although this reduces the difficulty of solving the upper-level problem, due to the smaller resolution of lower-level spatial, the features transmitted upwards are semantically precise but spatially ambiguous, thus the restoration ability of multi-scale network in spatial details is limited. Hence, there is an urgent need to improve the quality of restoration of high-frequency detail part. A simple approach is to introduce a frequency domain transform as an alternative to a spatial domain transform. This would provide the algorithm with a choice and direct its attention to different frequencies. 
In this paper, we consider using the discrete wavelet transform for the following reasons:
\begin{enumerate}
    \item Many recent state-of-the-art deblurring algorithms~\cite{mao2021deep,kong2023efficient,zou2021sdwnet} have introduced the discrete Fourier transform (DFT) as a frequency prior, which provides information that helps the algorithm to identify and select high-and low-frequency components that need to be preserved during restoration. Compared to DFT, the discrete wavelet transform (DWT) is better suited to deal with images containing more abrupt signals~\cite{daubechies1990wavelet}.
    \item Realistic blur and synthetic blur have significant distributional differences~\cite{rim2022realistic}. In the real world, due to the short exposure time of a camera, realistic blur has a specific directionality~\cite{tsai2022banet}, i.e., the blur trajectory is regionally continuous; conversely, synthetic blur has an unnatural and discontinuous trajectory. To fully utilize the potential deblurring guidance brought by this trajectory continuity, we use 2D-DWT to reveal blur directionality, distinguish changes in the blur signal along different directions, and provide the algorithm with a reliable basis for deblurring through adaptive learning.
\end{enumerate}

To make 2D-DWT fit the data distribution and feature layer space more closely, we implemented 2D-DWT with an adaptive data distribution by using group convolution, transferring feature space from the spatial domain into the wavelet domain, generating sub-signals with different frequency features and different directional features, and then constructing wavelet losses for them in order to constrain them by self-supervision. Experimental results demonstrate that the proposed method has advanced performance in terms of accuracy and efficiency (Fig.~\ref{fig:metrics}).

In summary, our contributions are as follows:
\begin{itemize}
   \item We propose a single-input and multiple-output(SIMO) multi-scale baseline for progressive image deblur, which reduces the complexity of existing multi-scale deblurring algorithms and improves the overall efficiency of image restoration networks.
   \item We construct a learnable discrete wavelet transform node(LWN) to provide reliable frequency and direction selection for the algorithm, which promotes the algorithm's restoration of the high-frequency components of edges and details.
   \item For the network to work better, reasonable multi-scale loss is proposed to guide pixel-by-pixel and scale-by-scale restoration. We also created reasonable self-supervised losses for the learnable wavelet transform to limit the learning direction of the wavelet kernel.
   \item We demonstrate the effectiveness of our algorithm under several motion blur datasets, especially real datasets, and obtain highly competitive results.
\end{itemize}

\begin{figure*}[htbp]  
\setlength{\abovecaptionskip}{+0.15cm}
\setlength{\belowcaptionskip}{-0.6cm}
    \centering
    \includegraphics[width=1.\linewidth]{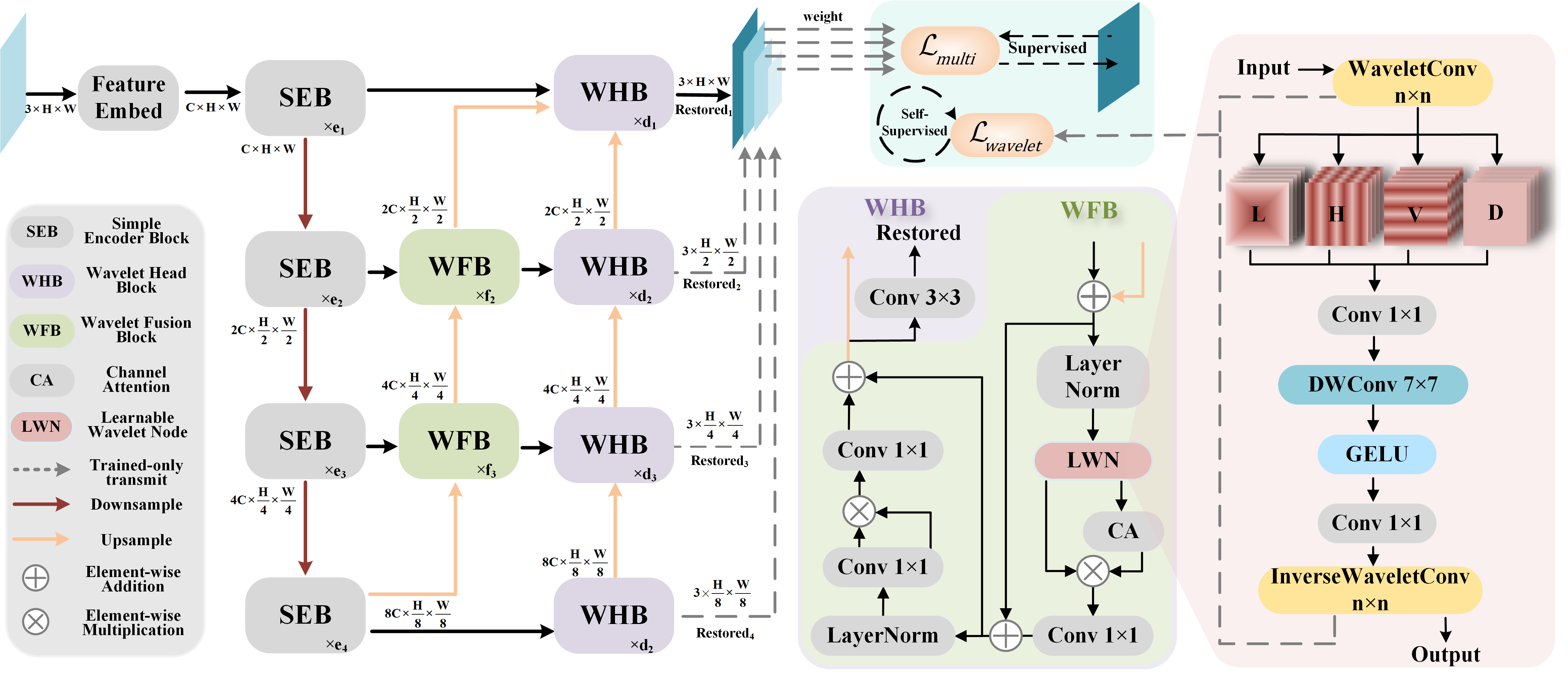}
    \caption{The overall architecture of the proposed MLWNet, the SEB is a simple module designed with reference~\cite{chen2022simple}, the WFB and WHB apply the LWN that implements the learnable 2D-DWT. In training phase, supervised learning is performed using $\mathcal{L}_{multi}$ and self-supervised restraint of the wavelet kernel is performed using $\mathcal{L}_{wavelet}$. In testing phase, only the highest scale restored images is output.}
    \label{fig:overall}
\end{figure*}

\section{Related Work}
\textbf{Single-scale Deblurring Algorithm.}
Image deblurring have developed rapidly in recent years~\cite{kupyn2018deblurgan,zhang2020deblurring,zhang2019deep,chen2022simple,kong2023efficient,li2023efficient}. To preserve richer image details, Kupyn et al.~\cite{kupyn2018deblurgan} viewed deblurring as a special case of image-to-image conversion, and for the first time utilized a GAN to recover images with richer details. To provide a high-quality and simple baseline, Chen et al.~\cite{chen2022simple} proposed a nonlinear activation-free network that obtained excellent results. Zamir et al.~\cite{zamir2022restormer} proposed a transformer with an encoder–decoder architecture that can be applied to higher-resolution images and utilizes cross-channel attention. Kong et al.~\cite{kong2023efficient} utilized a domain-based self-attention solver to reduce the transformer complexity and suppress artefacts. However, such single-scale algorithms estimate complex restoration problems directly and may require the design of restorers of higher complexity, which is suboptimal in terms of efficiency~\cite{kim2022mssnet}.

\noindent\textbf{Multi-scale Deblurring Algorithm.}
Multi-scale motion deblur methods aim to achieve progressive recovery using a multi-stage or multi-sub-network approach. The multi-scale algorithm previously used in the field of image deblurring is based on MIMO~\cite{nah2017deep,tao2018scale,park2020multi,cho2021rethinking,li2022learning}. Recently, Cho et al.~\cite{cho2021rethinking} proposed MIMO-UNet, which employs a multi-output single decoder as well as a multi-input single encoder to simulate a multi-scale architecture consisting of stacked networks, thereby greatly simplifying complexity. Zamir et al.~\cite{zamir2021multi} designed a multi-stage progressive image recovery network to simplify the information flow between multiple sub-networks. As shown in Sec.~\ref{baseline}, our proposed multi-scale approach further simplifies the structure of multi-scale networks and better balances the accuracy and complexity.

\noindent\textbf{Application of Frequency in Deblurring.}
Frequency is an important property of images, in recent years frequency domain has been introduced into numerous DNNs to be solved for various tasks~\cite{suvorov2022resolution,wang2022image,zhong2022detecting,kong2023efficient,yang2020fda,qin2021fcanet}. Mao et al.~\cite{mao2021deep} proposed DeepRFT, which introduced a residual module based on the Fourier transform into an advanced algorithm. Kong et al.~\cite{kong2023efficient} proposed FFTformer, which introduced the Fourier transform into a feed-forward network to effectively utilize different frequency information. Zou et al.~\cite{zou2021sdwnet} proposed SDWNet, which utilized frequency domain information as a complement to spatial domain information. Different from these methods, we introduce a learnable 2D-DWT capable of adapting to data distribution and feature space, more suitable not only for dealing with digital images rich in mutation signals, but also for dealing with real blurring.
\section{Proposed Method}
Our proposed MLWNet(Fig.~\ref{fig:overall}) aims to explore an efficient multi-scale architectural approach to achieve high-quality blind deblurring of a single image. First, we designed a novel and highly scalable SIMO multi-scale baseline to address the performance bottleneck faced by partially multi-scale networks. It takes a single image as an input and gradually generates a series of sharp images from the bottom up. Then, we propose a learnable wavelet transform node (LWN) for image deblurring, and it enhances the proposed algorithm’s ability to restore detail information.

\subsection{Multi-scale Baseline}
\label{baseline}
Our multi-scale network maintains an encoder-decoder architecture, but retains the original image with the highest resolution as input, and restores sharp images of various scales sequentially in the output stage. In this way, we eliminate the complexity of the fusion module when dealing RGB images of different scales and with deep semantics. 

For ease of understanding, we will introduce here the Encoder phase, multi-scale semantic Fusion phase (hereafter referred to as Fusion phase), and Decoder phase sequentially. 
As shown in Fig.~\ref{fig:overall}, the Encoder stage consists of several gray Simple Encoder Blocks(SEB), where information flows top-down for feature extraction; 
the Fusion stage consists of several green Wavelet Fusion Blocks(WFB), where information flows bottom-up for semantic fusion at different scales; 
and the Decoder stage consists of several purple Wavelet Head Blocks(WHB), and the bottom-up flow of information enables gradual restoration.
Compared with WHB, SEB replaces LWN with a $1\times1$ conv for channel scaling, and a $3\times 3$ depth-conv for feature extraction.

The Encoder stage is responsible for full feature extraction, downsampling feature maps once after each block, and then passing the feature maps to the Fusion stage or the Decoder stage as needed, whose outputs $E_{out}^i$ to block of the $i-th$ layer can be denoted as:
\begin{equation}
    \begin{aligned}
        E_{out}^{i} &= \left\{
            \begin{array}{lr}
                \phi_{i}\left(embed(x)\right), &i=i_{max} \\
                \phi_{i}\left(E_{out}^{i-1}\right), &otherwise \\
            \end{array}
            \right.\\
       \end{aligned}
     \label{eq:backbone}
 \end{equation}
where $\phi_{i}$ represent the block of corresponding layer of Encoder. The Fusion stage adapts and fuses information from different scales of semantics produced in Encoder to generate intermediate representations with deep semantics and shallow details. The Fusion stage's outputs $F_{out}^i$ to the block of $i-th$ layer is expressed as in Eq.~\ref{eq:neck}, where $\delta_{i}$ represents the block of the corresponding layer of  Fusion.
\begin{equation}
   \begin{aligned}
      F_{out}^{i} &= \left\{
         \begin{array}{lr}
            \delta_i\left(E_{out}^{i} + E_{out}^{i-1}\right), &i=i_{min} \\
            \delta_i\left(F_{out}^{i-1} + E_{out}^{i}\right), &otherwise \\
         \end{array}
         \right.\\
      \end{aligned}
	\label{eq:neck}
\end{equation}
The Decoder stage utilizes the information passed by both the Encoder and Fusion stages to progressively upsample the feature maps and generating pre-output feature maps at each scale separately. Output $D_{out}^i$ to the block of the $i-th$ layer is denoted as in Eq.~\ref{eq:head}. Here $\xi_{i}$ represents the block of the corresponding layer of the Decoder.
\begin{equation}
   \begin{aligned}
      D_{out}^{i} = 
         \begin{cases}
            \xi_i\left(E_{out}^{i}\right), \qquad \qquad \qquad \qquad \quad i=i_{min}\\
            \xi_i\left(D_{out}^{i-1} + F_{out}^{i}\right), \qquad i>i_{min}\text{ and }i<i_{max} \\
            \xi_i\left(D_{out}^{i-1} + F_{out}^{i-1}+E_{out}^i\right), \qquad i=i_{max}
         \end{cases} \nonumber
      \end{aligned}
   \label{eq:head}
\end{equation}

Each block in the Decoder is followed by a $3\times3$ convolution to generate a restored image at each scale. To improve inference efficiency, all but the highest-scale outputs are generated in the training phase to compute the multi-scale loss. We note that this design also have a similar role with auxiliary head~\cite{yu2021bisenet} in facilitating the model's learning of the recovered uniform patterns.

\begin{figure}[htbp]
\setlength{\abovecaptionskip}{-0.15cm}
\setlength{\belowcaptionskip}{-0.3cm}
    \begin{center}
    \includegraphics[width=1.\linewidth]{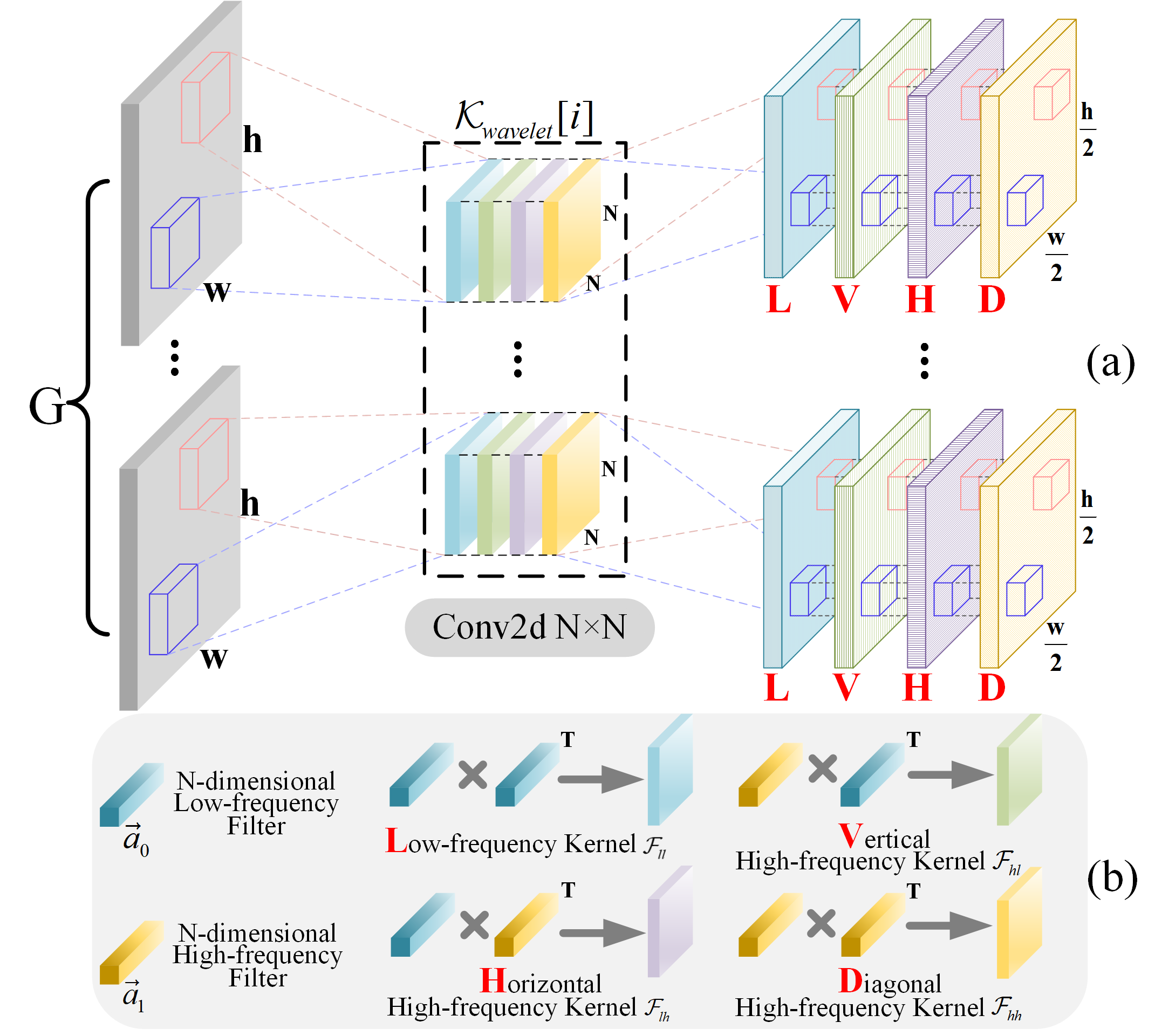}
    \end{center}
    \caption{(a)The process of learnable 2D-wavelet convolution. (b)The construction process of the $N \times N$ 2D-wavelet kernel.}
    \label{fig:wavelet-conv}
\end{figure}

\subsection{Learnable Discrete Wavelet Transform}
\label{ldwt}
To alleviate the defects of multi-scale architectures in the recovery of detail information, we design LWN and constructed the WFB and WHB based on it. Unlike the Fourier transform, the wavelet transform has adaptive time-frequency resolution, and performs excellently on digital images rich in mutation signal. Its use in conjunction with DNNs has gained increasing widespread attention~\cite{liu2020densely,liu2019multi,wolter2021adaptive,ha2021adaptive}.

For ease of understanding, the 1D-DWT case is given here first. For the input discrete signal $t$, given the wavelet function $\psi_{j, k}(t)=2^{\frac{j}{2}}\psi(2^{j}t -k)$ for scaling factor $j$, time factor $k$, and scale function $\phi_{j, k}(t)=2^{\frac{j}{2}}\phi( 2^{j}t -k)$, the decomposition of the input signal $t$ at $j_0$ is given by:
\begin{equation}
   f(t)=\sum_{j>j_{0}} \sum_k d_{j, k} \psi_{j, k}(t) + \sum_k c_{j_0,k}\phi_{j_0,k}\left(t\right)
\end{equation}
In this process, $d_{j,k}=\left\langle f(t), \psi_{j, k}(t)\right\rangle$ represents the detail coefficients (i.e., high-frequency components),
$c_{j_0,k}=\left\langle f(t), \phi_{j_0, k}(t)\right\rangle$ represents the approximation coefficients (i.e., low-frequency components) so that the input signal can be disassembled step by step into a rich wavelet domain signal.
In order to use wavelet transform in DNN, we introduce the analysis vectors $\vec{a}_0[k]=\left\langle\frac{1}{\sqrt{2}} \phi(\frac{t}{2}), \phi(t-k)\right\rangle$, $\vec{a}_1[k]=\left\langle\frac{1}{\sqrt{2}} \psi(\frac{t}{2}), \phi(t-k)\right\rangle$ to represent the high and low frequency filters respectively, 
then:
\begin{equation}
   \begin{aligned}
      &c_{j+1,p}=\sum_k \vec{a}_0[k-2p]c_{j,k}\\
      &d_{j+1,p}=\sum_k \vec{a}_1[k-2p]c_{j,k}
   \end{aligned}
   \label{eq:cd}
\end{equation}
\par According to Eq.~\ref{eq:cd}, the decomposition of the original signal over the wavelet basis can be viewed as a recursive convolution of the original signal with a specific filter using a step size of 2. Similarly, the inverse wavelet transform can be realized by transposed convolution of two synthetic vectors $\vec{s}_0$ and $\vec{s}_{1}$. 
We extend the above situation to 2D by efficiently extracting axial high-frequency information using 2D discrete wavelet transform (2D-DWT), and designing learnable methods to make it adaptive to the data distribution and feature layers. 
Fig.~\ref{fig:wavelet-conv}(b) presents the construction process of the forward wavelet convolution kernel in 2D, which can be expressed as:
\begin{equation}
   \begin{aligned}
   \mathcal{F}_{ll} &= \vec{a_0} \times \vec{a_0}^{ T}, \mathcal{F}_{lh} = \vec{a_0} \times\vec{a_1}^{ T}\\
   \mathcal{F}_{hl} &= \vec{a_1} \times \vec{a_0}^{ T}, \mathcal{F}_{hh} = \vec{a_1} \times \vec{a_1}^{ T}\\
   \mathcal{K}_{w} &= \text{cat}\left(\mathcal{F}_{ll}, \mathcal{F}_{lh},\mathcal{F}_{hl},\mathcal{F}_{hh}\right)
   \end{aligned}
\end{equation}
\par We set $\vec{a_0}$ and $\vec{a_1}$ as learnable filters; 
$\mathcal{F}_{ll}$, $\mathcal{F}_{lh}$, $\mathcal{F}_{hl}$, and $\mathcal{F}_{hh}$ are low-frequency, horizontal high-frequency, vertical high-frequency, and diagonal high-frequency convolution operators obtained from vector outer products; The wavelet kernel $\mathcal{K}_{w}$ is spliced from above convolution operators.
\par Then we implement learnable wavelet transform as a group convolution like Fig.~\ref{fig:wavelet-conv}(a), given a set of input feature maps $\mathcal{X}_{in} \in \left(C, H, W\right)$, its projection in wavelet domain $\mathcal{X}_{out} \in \left(4C, \frac{H}{2}, \frac{W}{2}\right)$ (low-frequency, horizontal high-frequency, vertical high-frequency, and diagonal high-frequency component) is generated through the wavelet convolution. 
Similarly, the construction method of the inverse kernel and the forward propagation can be easily deduced, since they are inverse processes of each other.
\par Then, a natural question can then be posed: how to ensure the correctness of the wavelet kernel learning and ensure that it does not degrade into a general group convolution and cause performance degradation?
\par A common practice is to introduce the principle of perfect reconstruction~\cite{wolter2021adaptive,liu2019multi} to constrain adaptive wavelets, for a complex $z \in \mathbb{C}$, given a filter $\vec{x}$ that applies to the $z$-transform, its $z$-transform can be expressed as $X(z) = \sum_{n \in \mathbb{Z}} \vec{x}(n) z^{-n} $. 
Then $\vec{a}_0$,$\vec{a}_1$,$\vec{s}_0$,$\vec{s}_1$ can be obtained as their corresponding z-transforms $A_{0}$,$A_{1}$,$S_{0}$,$S_1$, which must then be satisfied if perfect reconstruction is desired:
\begin{equation}
	\begin{aligned}
	A_0(-z)S_0(z) + A_1(-z)S_1(z) = 0
	\end{aligned}
	\label{eq:ac}
\end{equation}
\begin{equation}
	\begin{aligned}
	A_0(z)S_0(z) + A_1(z)S_1(z) = 2
	\end{aligned}
	\label{eq:pr}
\end{equation}
\par Eq.~\ref{eq:ac} is known as alaising cancellation condition, which is used to the cancellation of the aliasing effects arising from the downsampling.
We will give a detailed derivation of Eq.~\ref{eq:ac} and Eq.~\ref{eq:pr} in the Appendix.
\par After wavelet convolution, to achieve frequency restoration in the wavelet domain, the input $\mathcal{X}$ wavelet domain component was separated into a separate dim, we use depth-wise convolution with an expansion factor of $r$ for wavelet domain feature extraction and transformation after the wavelet positive transform, and $1\times1$ convolution for channel expansion and scaling, and finally a learnable wavelet inverse transform is used to reduce the wavelet domain feature maps to the spatial domain for output, which constitutes the LWN. We follow the rules of~\cite{chen2022simple} and design structurally similar Wavelet Head Block (WHB) and Wavelet Fusion Block (WFB), both with LWN as an important base module, but used for semantic aggregation and multiscale output at different scales, respectively; 
thus, WHB adds the recovery convolution branch after WFB.

\subsection{Loss}
\label{loss}
\textbf{Multi-scale Loss}. Follow PSNR loss~\cite{chen2022simple}, we propose multi-scale loss function as the main loss of the algorithm for calculating the pixel difference between restored image and GT at each scale:
\begin{equation}
	\begin{aligned}
		\mathcal{L}_{multi}(x, y) &= \sum_{k=1}^{K} w_k \times \mathcal{L}_{psnr}\left(x_k, y_k\right) \\
	\end{aligned}
	\label{eq:multi_loss}
\end{equation}
where $x$, $y$ denote the output and GT, respectively, $k$ denotes the downsampling level, $w_k$ represents the weights under the corresponding scale.
Our design of $w_{k}$ is based on the simple finding that we cannot produce absolutely accurate downsampled clear images, 
and that the accuracy gap is progressively enlarged with decreasing scales.
To ensure that lower scale output layers do not negatively affect the final output, we empirically set the loss weight $w_k$ to $\frac{1}{k}$ for the corresponding k-scale.

\noindent\textbf{Wavelet Loss}. Both Eq.~\ref{eq:ac} and Eq.~\ref{eq:pr} can be easily converted losses optimized toward a minimum of 0, here we use the mean square error.
Then, based on the convolvable nature of the $z$ transform, it is possible to construct convolutionally equivalent substitutions for polynomial multiplications for the perfect reconstruction condition:
\begin{equation}
   \begin{aligned}
   L_{wavelet}(\theta_i)&=\left(\sum_{k}^{N-1}\left(\left \langle a_0, s_0 \right \rangle_{k} + \left \langle a_1, s_1\right \rangle_{k} \right) - \hat{V}_{\lfloor \frac{N}{2} \rfloor }\right)^2
   \\
   &=\left(\sum_{k}^{N-1}\left(\left \langle \left(-1\right)^ka_0, s_0 \right \rangle_{k} + \left \langle (-1)^ka_1, s_1\right \rangle_{k} \right)\right)^2
   \end{aligned}
   \label{eq:ac_loss}
\end{equation}
where $k$ denotes the position of the filter when it performs the convolution, $\hat{V}$ is a vector with a center position value of two, and $\theta_i$ denotes the constructed filter

\noindent\textbf{Overall Loss}.Based on the multi-scale loss and wavelet loss, the loss function used in this study is:
\begin{equation}
	\mathcal{L}_{total}(x, y) = \mathcal{L}_{wavelet}(\theta_i) + \mathcal{L}_{multi}(x, y)
	\label{eq:total_loss}
\end{equation}
\section{Experimental Results}
\subsection{Datasets and Implementation Details}
We evaluate our method on the realistic datasets RealBlur~\cite{rim2020real} and RSBlur~\cite{rim2022realistic}, as well as the synthetic dataset GoPro~\cite{nah2017deep}, they are all composed of blurry-sharp image pairs. During training we use AdamW optimizer ($\alpha=0.9$ and $\beta=0.9$) for a total of 600K iterations. The initial value of the learning rate is $10^{-3}$ and is updated with cosine annealing schedule. The patch size is set to 256 × 256 pixels and we followed the training strategy of~\cite{zamir2022restormer,tsai2022stripformer}. For data augmentation, we only use random flips and rotations. More experimental results can be found in the Appendix.
\begin{figure*}[htbp]
   \setlength{\abovecaptionskip}{-0.3cm}
   \setlength{\belowcaptionskip}{-0.3cm}
       \begin{center}
       \includegraphics[width=1.\linewidth]{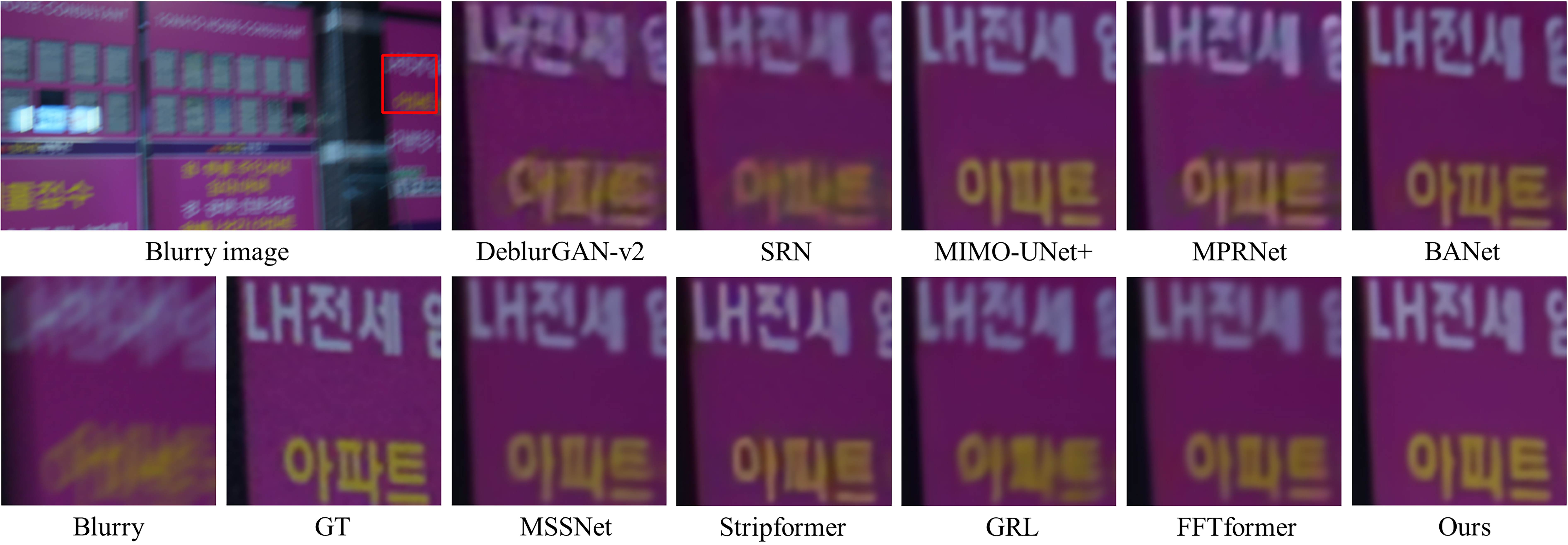}
       \end{center}
       \caption{Visual comparisons on the RealBlur-J dataset~\cite{rim2020real}. The proposed method generates an image with clearer characters.}
       \label{fig:result_realblur}
   \end{figure*}
\subsection{Comparison with State-of-the-Art Methods}
We have compared our method with several advanced deblurring methods and use the PSNR and SSIM to evaluate the quality of restored images.

\begin{table}[htp]
    \centering
\setlength{\abovecaptionskip}{+0.1cm}
\setlength{\belowcaptionskip}{-0.7cm}
	\setlength{\tabcolsep}{2pt}
    \begin{tabular}{l|c|c|c|c||c}
       \toprule
       \multirow{2}{*}{Method} & \multicolumn{2}{c}{RealBlur-J} & \multicolumn{2}{c}{RealBlur-R} & \multirow{2}{*}{\makecell[c]{Avg.\\runtime}} \\
       & PSNR  & SSIM & PSNR  & SSIM & \\
       \midrule
       DeblurGAN-v2~\cite{kupyn2019deblurgan} & 29.69 & 0.870 & 36.44 & 0.935 & 0.04s\\
       SRN~\cite{tao2018scale}         & 31.38 & 0.909 & 38.65 & 0.965 & 0.07s\\
       MPRNet~\cite{zamir2021multi}       & 31.76 & 0.922 & 39.31 & 0.972 & 0.09s\\
       SDWNet~\cite{zou2021sdwnet}       & 30.73 & 0.896 & 38.21 & 0.963 & 0.04s \\
       MIMO-UNet+~\cite{cho2021rethinking}  & 31.92 & 0.919 & - & - & 0.02s\\
       MIMO-UNet++~\cite{cho2021rethinking} & 32.05 & 0.921 &  - & -  & -\\
       DeepRFT+~\cite{mao2021deep}     & 32.19 & 0.931 & 39.84 & 0.972 & 0.09s\\
       BANet~\cite{tsai2022banet}      & 32.00 & 0.923 & 39.55 & 0.971 & 0.06s\\
       BANet+~\cite{tsai2022banet}     & 32.42 & 0.929 & 39.90 & 0.972 & 0.12s\\
       Stripformer~\cite{tsai2022stripformer}  & 32.48 & 0.929 & 39.84 & \underline{0.974} & 0.04s \\
       MSSNet~\cite{kim2022mssnet}      & 32.10 & 0.928 & 39.76 & 0.972 & 0.06s\\
       MSDI-Net~\cite{li2022learning}     & 32.35 & 0.923 &  -  &  -  & 0.06s\\
       MAXIM-3S~\cite{tu2022maxim}   & 32.84 & \underline{0.935} & 39.45 & 0.962 &  -\\
       FFTformer~\cite{kong2023efficient}  & 32.62 & 0.933 & 40.11 & 0.973 & 0.13s\\
       GRL-B~\cite{li2023efficient}       & 32.82 & 0.932 & \underline{40.20} & \underline{0.974} & 1.28s\\
       \midrule
       MLWNet-S    & \underline{33.02} & 0.933  &    -   &     -   & 0.04s \\
       MLWNet-B    & \textbf{33.84} & \textbf{0.941} & \textbf{40.69} & \textbf{0.976} & 0.05s\\
       \bottomrule
    \end{tabular}
    \caption{Quantitative evaluations on the RealBlur dataset~\cite{rim2020real}. 
    The experimental results were trained under the corresponding datasets respectively, and average runtime is tested on 256$\times$256 patchs.}
    \label{tab:RealBlur}
\end{table}

\begin{table}[htp]
   \centering
\setlength{\abovecaptionskip}{+0.1cm}
\setlength{\belowcaptionskip}{-0.2cm}
  \setlength{\tabcolsep}{6.5pt}
  \begin{tabular}{l|c|c|c|c}
      \toprule
      \multirow{2}{*}{Method} & \multicolumn{2}{c}{RSBlur} & \multicolumn{2}{c}{RealBlur-J} \\
      & PSNR  & SSIM & PSNR  & SSIM \\
      \midrule
      SRN~\cite{tao2018scale}         & 32.53 & 0.840 & 29.86 & 0.886 \\
      MIMO-Unet~\cite{cho2021rethinking}    & 32.73 & 0.846 & 29.53 & 0.876 \\
      MIMO-Unet+~\cite{cho2021rethinking}  & 33.37 & 0.856 & 29.99 & 0.889 \\
      MPRNet~\cite{zamir2021multi}      & 33.61 & 0.861 & 30.46 & \underline{0.899} \\
      Restormer~\cite{zamir2022restormer}    & 33.69 & 0.863 & \underline{30.48} & 0.891 \\
      Uformer-B~\cite{wang2022uformer}   & 33.98 & 0.866 & 30.37 & \underline{0.899} \\
      SFNet~\cite{cui2023selective}       & \underline{34.35} & \underline{0.872} & 30.26 & 0.897 \\
      \midrule
      MLWNet-B     & \textbf{34.94} & \textbf{0.880} & \textbf{30.53} & \textbf{0.905} \\
      \bottomrule
   \end{tabular}
   \caption{Quantitative evaluations trained on the RSBlur dataset~\cite{rim2022realistic}, the RealBlur-J dataset was used for testing only.}
   \label{tab:RSBlur}
\end{table}

\begin{table}[htp]
   \centering
\setlength{\abovecaptionskip}{+0.1cm}
\setlength{\belowcaptionskip}{-0.55cm}
  \setlength{\tabcolsep}{7pt}
   \begin{tabular}{l|c|c||c}
      \toprule
      \multirow{2}{*}{Method} & \multicolumn{2}{c}{~\cite{rim2020real}$\rightarrow$ ~\cite{rim2022realistic}} & \multirow{2}{*}{\makecell[c]{MACs(G)}} \\
       & PSNR  & SSIM & \\
      \midrule
      DeblurGAN-v2~\cite{kupyn2019deblurgan}  & 30.15 & 0.766 & 42.0 \\
      MPRNet~\cite{zamir2021multi}       & 29.56 & 0.785 & 760.8  \\
      MIMO-UNet+~\cite{cho2021rethinking}   & 29.69 & 0.792 & 154.4 \\
      BANet~\cite{tsai2022banet}       & 30.19 & 0.806 & 263.9\\
      BANet+~\cite{tsai2022banet}      & \underline{30.24} & \underline{0.809} & 588.7 \\
      MSSNet~\cite{kim2022mssnet}       & 29.86 & 0.806 & 154.0\\
      FFTformer~\cite{kong2023efficient}   & 29.70 & 0.787 & 131.8\\
      \midrule
      MLWNet-B     & \textbf{30.91} & \textbf{0.818} & 108.2 \\
      \bottomrule
   \end{tabular}
   \caption{Quantitative evaluation for generalizability shows the results of models trained on the RealBlur-J dataset and tested on the RSBlur dataset,
   MACs are measured on 256 × 256 patches.}
   \label{tab:J2RSBlur}
\end{table}
\noindent\textbf{Evaluations on the RealBlur dataset.} Since our proposed method is driven by real-world deblurring, we first conduct comparisons on RealBlur~\cite{rim2020real}. The quantitative analysis results are shown in Tab.~\ref{tab:RealBlur}, the PSNR value of our method is 0.91dB and 0.49dB higher than state-of-the-art GRL on the RealBlur-J and RealBlur-R datasets respectively. Compared to other multi-scale architectures, our method has fewer model computational and running time while the performance is better. Fig.~\ref{fig:result_realblur} shows visual comparison on RealBlur-J dataset, our method obtains restored text with richer edge details, while achieving the best contrast, sharpness, brightness, and structural details of the image.

\begin{figure*}[htbp]
\setlength{\abovecaptionskip}{-0.3cm}
\setlength{\belowcaptionskip}{-0.45cm}
    \begin{center}
    \includegraphics[width=1.\linewidth]{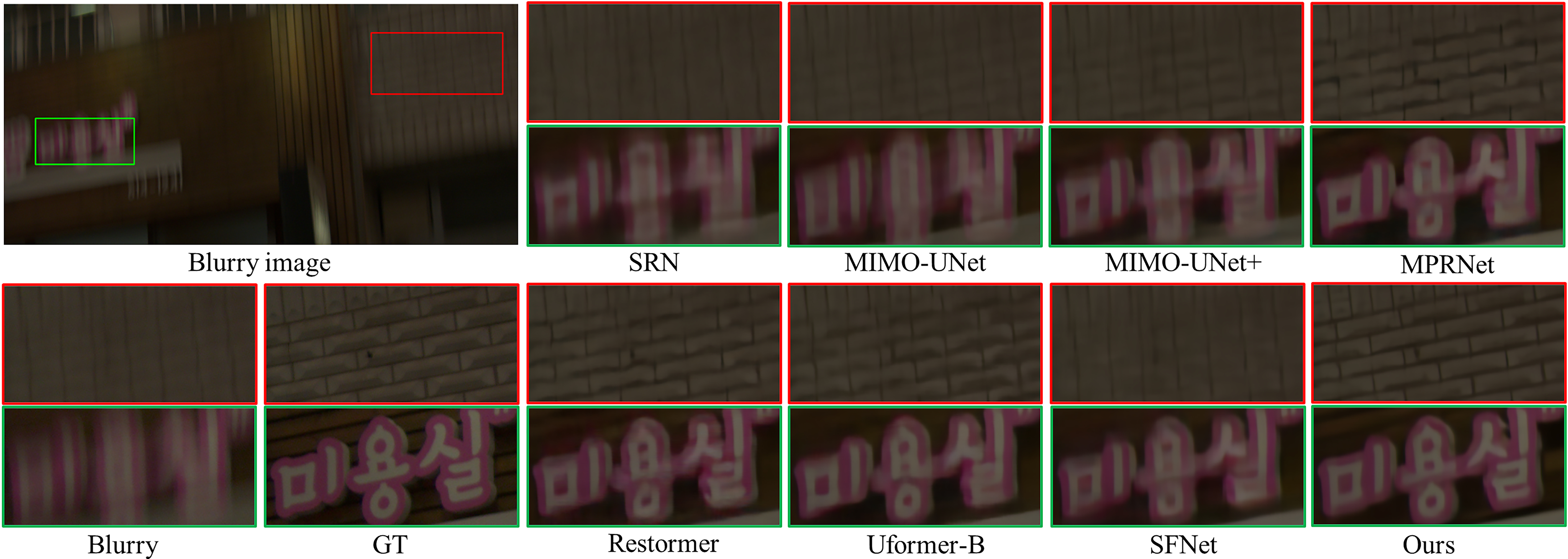}
    \end{center}
    \caption{Visual comparisons on the RSBlur dataset~\cite{rim2022realistic}. The deblurring performance of the proposed method in low-light is impressive, The recovery of characters and texture structures far exceeds other advanced methods.}
    \label{fig:result_rsblur}
\end{figure*}

\noindent\textbf{Evaluations on the RSBlur dataset.} We further conduct experiments on the latest real-world deblurring dataset RSBlur~\cite{rim2022realistic} and follow the protocol of this dataset for fair comparison. Tab.~\ref{tab:RSBlur} summarizes the quantitative evaluation results of our method with advanced algorithms. The proposed MLWNet achieved the highest PSNR and SSIM, which were 34.94 and 0.880 respectively. In addition, the model we trained on RSBlur dataset also achieved the best results on RealBlur-J dataset. We show some visual comparisons in Fig.~\ref{fig:result_rsblur}. We note that our method outperforms other methods in low-light blurry scenes, proving that the proposed method makes textures and structures clearer.

In addition, we evaluate the RSBlur dataset using models trained only on RealBlur-J to fairly compare the generalization of methods to real scenarios. As shown in Tab.~\ref{tab:J2RSBlur}, our method produced results with highest PSNR value of 30.91. The results of Tab.~\ref{tab:RSBlur} and Tab.~\ref{tab:J2RSBlur} show that our model has a better generalization ability.

\begin{figure*}[htbp]
\setlength{\abovecaptionskip}{-0.3cm}
\setlength{\belowcaptionskip}{-0.45cm}
    \begin{center}
    \includegraphics[width=1.\linewidth]{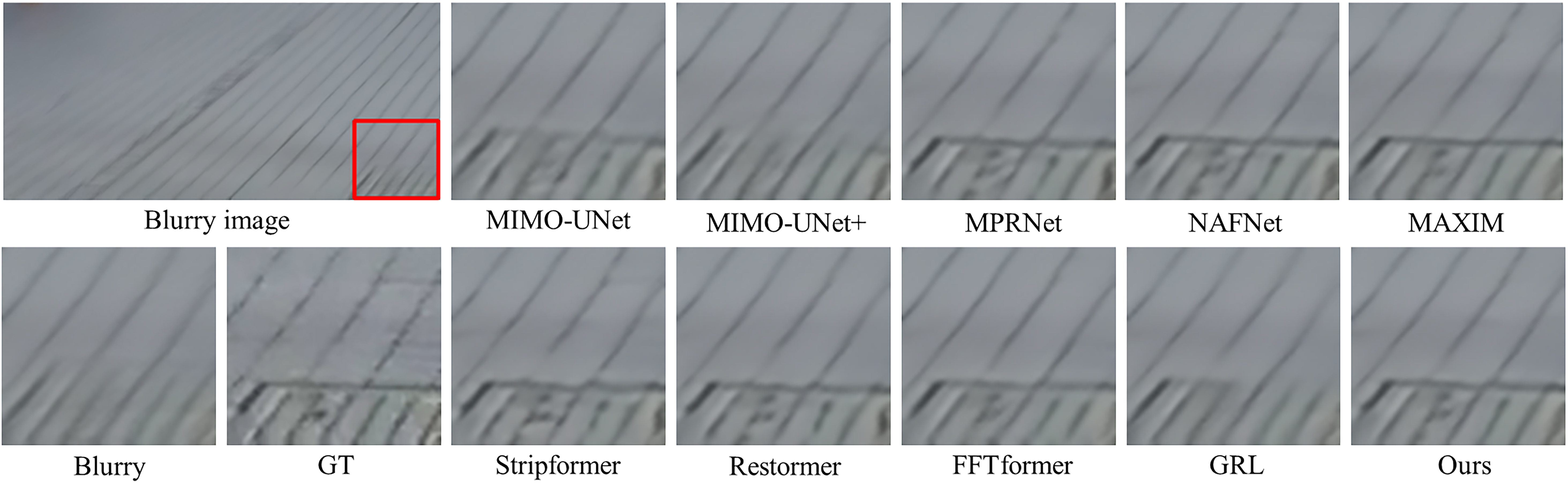}
    \end{center}
    \caption{Visual comparisons on the GoPro dataset~\cite{nah2017deep}. Our method better preserves texture information without sharpening.}
    \label{fig:result_gopro}
\end{figure*}

\noindent\textbf{Evaluations on the GoPro dataset.} We conduct a extensive comparison with advanced algorithms on GoPro~\cite{nah2017deep}. Tab.~\ref{tab:gopro} shows the quantitative evaluation results. The gap between our method and the state-of-the-art FFTformer is only 0.001 in SSIM value, while the running time is reduced by 60\%. We show a visual comparison on the GoPro dataset in Fig.~\ref{fig:result_gopro}. We can see that our method recovers blurred textures well without sharpening. For the reasons why the proposed MLWNet does not achieve optimal performance in synthetic blurry, we conducted a detailed analysis in Sec.~\ref{analysis}.

\begin{table}[htp]
   \centering
\setlength{\abovecaptionskip}{+0.1cm}
\setlength{\belowcaptionskip}{-0.7cm}
	\setlength{\tabcolsep}{6.5pt}
	\begin{tabular}{l|c|c||c}
      \toprule
      \multirow{2}{*}{Method} & \multicolumn{2}{c}{GOPRO} & \multirow{2}{*}{\makecell[c]{Avg.runtime}}\\
      & PSNR  & SSIM &  \\
      \midrule
      DeblurGAN-v2~\cite{kupyn2019deblurgan}  & 29.55 & 0.934 & 0.04s \\
      SRN~\cite{tao2018scale}           & 30.26 & 0.934 & 0.07s \\
      DMPHN~\cite{zhang2019deep}         & 31.20 & 0.945 & 0.21s \\
      SDWNet~\cite{zou2021sdwnet}        & 31.26 & 0.966 & 0.04s\\
      MPRNet~\cite{zamir2021multi}        & 32.66 & 0.959 & 0.09s \\
      MIMO-UNet+~\cite{cho2021rethinking}    & 32.45 & 0.957 & 0.02s \\
      DeepRFT+~\cite{mao2021deep}      & 33.23 & 0.963 & 0.09s \\
      MAXIM-3S~\cite{tu2022maxim}      & 32.86 & 0.961 & -\\
      Stripformer~\cite{tsai2022stripformer}   & 33.08 & 0.962 & 0.04s \\
      MSDI-net~\cite{li2022learning}      & 33.28 & 0.964 & 0.06s \\
      Restormer~\cite{zamir2022restormer}     & 33.57 & 0.966 & 0.08s \\
      NAFNet~\cite{chen2022simple}        & 33.69 & 0.967 & 0.04s \\
      FFTformer~\cite{kong2023efficient}     & \textbf{34.21} & \textbf{0.969} & 0.13s \\
      GRL-B~\cite{li2023efficient}         & \underline{33.93} & \underline{0.968} & 1.28s\\
      \midrule
      MLWNet-B       & 33.83 & \underline{0.968} & 0.05s \\
      \bottomrule
   \end{tabular}
   \caption{Quantitative evaluations trained and tested on the GoPro dataset~\cite{nah2017deep}. Our proposed MLWNet obtains competitive results with a combination of time efficiency and accuracy.}
   \label{tab:gopro}
\end{table}

\section{Analysis and Discussion}
\label{analysis}
In this section, we provide a more in-depth analysis and show the contribution of each component of the proposed method.
We perform 20w iterations on the RealBlur-J dataset for abtation with a batch size of 8 to train our method using a version of the model with a width of 32, and the results are shown in Tab.~\ref{tab:wavelet}.

\noindent\textbf{Single-scale vs. Multi-scale}. 
Given that each node of the baseline network is composed of SEB, we have deformed the multi-scale and single-scale according to the architecture.
Tab.~\ref{tab:sm} shows that the multi-scale architecture improves the PSNR and SSIM values to varying degrees, proving that the SIMO strategy we employ helps to eliminate motion blur to some extent.
As shown in Sec.~\ref{baseline}, we use SIMO for coarse-to-fine restoration, and only use multiple outputs to calculate multi-scale losses during training.

\noindent\textbf{WFB and WHB}.
We default to using WFB and WHB via $\mathcal{L}_{wavelet}$ statute in this section, we also proved the effectiveness of LWN, because LWN is a subpart of the first two.
After applying them, the PSNR index improved by 0.25dB compared to the multi-scale baseline, which shows that LWN and its extension modules can easily help multi-scale algorithms improve detail recovery capabilities. 
Fig.~\ref{fig:feature} shows the four types of feature maps generated after forward learnable wavelet convolution. We can see that the high and low frequency information of the input feature map are mixed together, and after wavelet group convolution, the network exerts different attention on the feature information in different directions or frequencies.

\noindent\textbf{Effectiveness of $\mathcal{L}_{wavelet}$}.
By adding unreduced WFB and WHB to the baseline without using wavelet loss, the performance improvement of the baseline is minimal and does not break the bottleneck of multiscale detail recovery, which suggests that the wavelet convolution degenerates into a general group convolution that only serves to deepen the network, proving that merely deepening the network is ineffective for the restoration of detail information.
\begin{table}[htp]
   \vspace{-0.2cm}
   \renewcommand\arraystretch{0.95}
       \centering
      \setlength{\tabcolsep}{5.5pt}
      \begin{tabular}{lccc}
          \hline
          Method & SISO &  MIMO & SIMO   \\
          \hline
         PSNR/SSIM & \underline{32.29}/0.924 & 32.19/\underline{0.928} & \textbf{32.37}/\textbf{0.929}  \\
         MACs(G) & \textbf{19.24} & 21.83 & \underline{19.29}  \\ \hline
       \end{tabular}
   \vspace{-0.3cm}
   \caption{Comparison in various input and output modes.}
   \label{tab:sm}
   \end{table}

\begin{table}[htbp]
   \vspace{-0.8cm}
   \centering
   \setlength{\tabcolsep}{5pt}
   \setlength{\abovecaptionskip}{+0.1cm}
   \setlength{\belowcaptionskip}{-0.7cm}
   \begin{tabular}{ccccccc}
      \toprule
      SIMO & \small{WFB} & \small{WHB} & \small{$\mathcal{L}_{wavelet}$} & \footnotesize{PSNR}  & \footnotesize{SSIM} & \footnotesize{MACs(G)}\\
      \midrule
      \checkmark &            &            &            & 32.37  & 0.929 & 19.29 \\
      \checkmark & \checkmark & \checkmark &            & 32.40  & 0.928 & 28.21 \\
      \checkmark &            & \checkmark & \checkmark & 32.49  & 0.928 & 25.28 \\ 
      \checkmark & \checkmark &            & \checkmark & 32.57  & 0.929 & 22.22 \\
      \checkmark & \checkmark & \checkmark & \checkmark & 32.62  & 0.931 & 28.21 \\
      \bottomrule
   \end{tabular}
   \caption{Ablation study on components of the proposed MLWNet. We set the baseline network to use SEB in its entirety, and models that do not use SIMO will represent single scales using SISO.}
   \label{tab:wavelet}
\end{table}

\begin{figure}[htbp]
\vspace{-0.2cm}   
\setlength{\abovecaptionskip}{-0.25cm}
\setlength{\belowcaptionskip}{-0.5cm}
    \begin{center}
    \includegraphics[width=1.\linewidth]{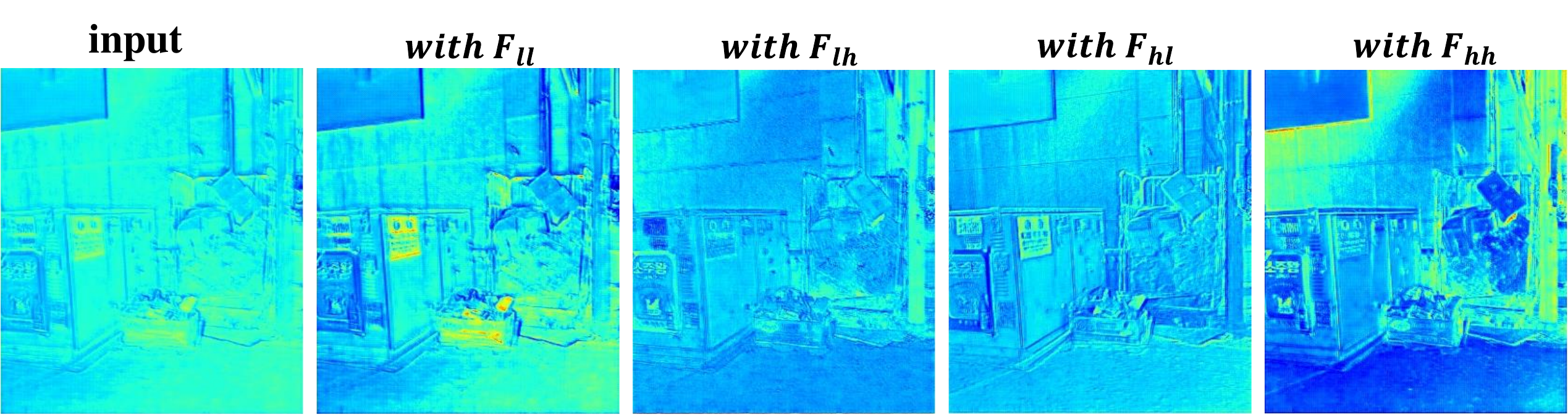}
    \end{center}
    \caption{Feature maps representing high-and low-frequency components generated after learnable wavelet convolution. Zoom in on the screen for the best view.}
    \label{fig:feature}
 \end{figure}

\begin{figure}[htbp]
\setlength{\abovecaptionskip}{-0.3cm}
\setlength{\belowcaptionskip}{-0.4cm}
    \begin{center}
    \includegraphics[width=1.\linewidth]{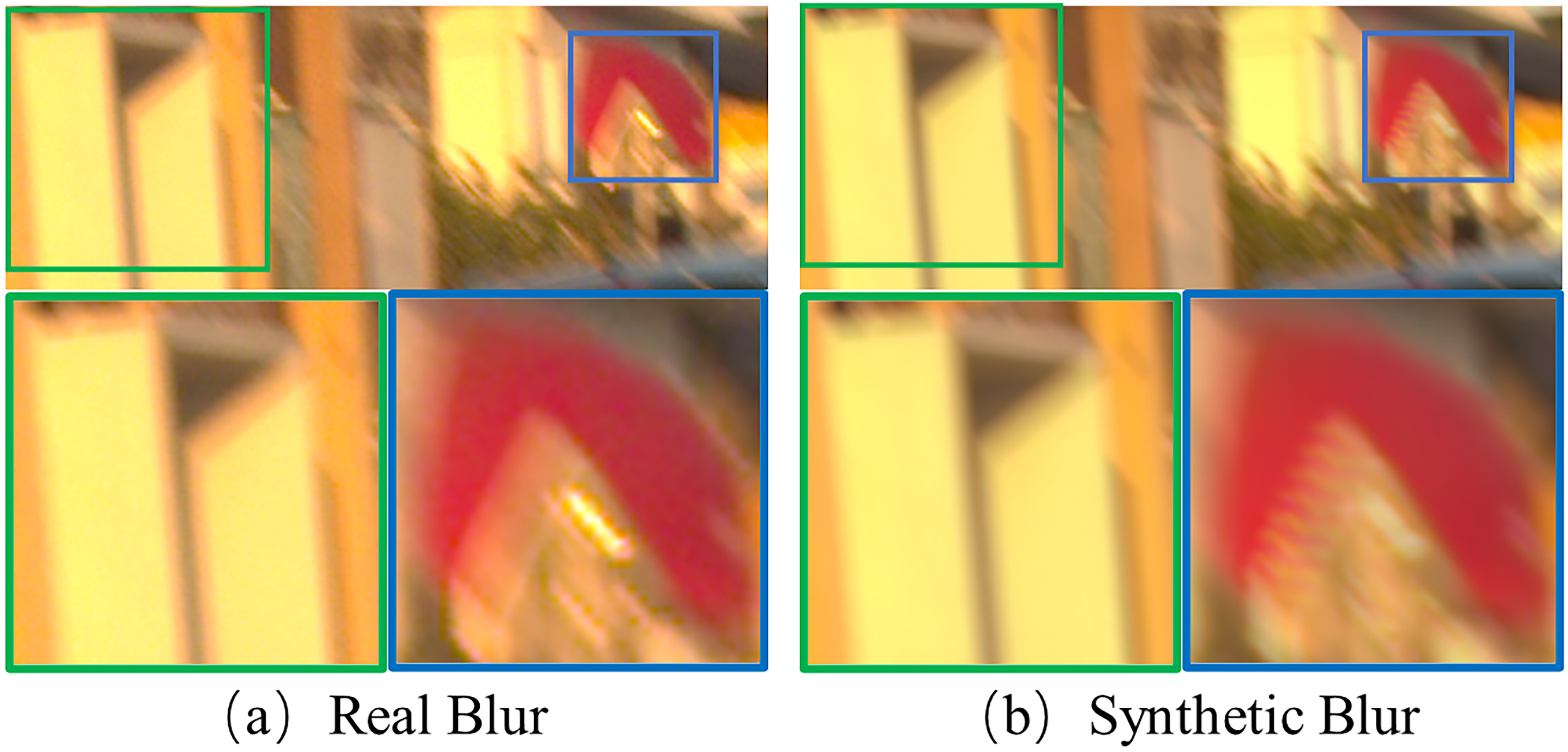}
    \end{center}
    \caption{The difference between realistic blur (a) and synthetic blur (b). In the green box, the synthetic blur appears with color averaging resulting in high and low frequency confusion, and in the blue box has unnatural discontinuous trajectories.}
    \label{fig:synthetic}
 \end{figure}

\noindent\textbf{Limitations and analysis}. It is observed from Tab.~\ref{tab:RealBlur}-\ref{tab:J2RSBlur} that our method fails to achieve the same expected high performance with synthetic blur as it does with realistic blur. We try to analyze the reasons for this phenomenon via dataset comparison and experimental results. First, the average frame synthesis method used in the synthetic blur dataset leads to unnatural discontinuous blur trajectories, thereby introducing strange high-frequency information interference (see Fig.~\ref{fig:synthetic}). Second, synthetic data tends to produce an unnatural mixture of high and low frequencies. It is easy to mix with the average color area at the edge of the texture, causing the confusion of high-low frequency information.

\vspace{-3mm}
\begin{table}[htbp]
\setlength{\abovecaptionskip}{+0.1cm}
\setlength{\belowcaptionskip}{-0.45cm}
   \centering
   \begin{tabular}{cccccc}
      \toprule
      Noise Level & L1 & L2 & L3 & L4 & L5 \\
      \midrule
      GoPro & 34.81 & 34.66 & 33.76 & 33.19 & 32.63 \\
      RealBlur-J & 33.92 & 33.81 & 33.97 & 33.93 & 33.54 \\
      \bottomrule
   \end{tabular}
   \caption{Performance comparison at different noise difference levels, where L3 contains the noise difference mean.}
   \label{tab:noise}
\end{table}

Finally, there is a large difference in noise levels between synthetic blur and real blur. We follow~\cite{chen2015efficient} to calculate the noise level of each testset, and divided the noise difference between clear images and blurred images of GoPro and RealBlur-J into 5 levels for separate testing. As shown in Tab.~\ref{tab:noise}, we note that as the noise difference increases, our method maintains good performance on RealBlur-J, while the performance drops fast on GoPro. This shows that our method is robust to the noise of real blur, and that there are differences in the noise between synthetic blur and real blur.

\section{Conclusion}
In this paper, we propose a SIMO-based multi-scale architecture to achieve efficient motion deblurring. We have developed a learnable discrete wavelet transform module, which not only improves the algorithm's ability to recover details, but is also more applicable to the real world. In addition, we construct a reasonable multi-scale loss to guide the recovery of blurred images pixel by pixel and scale by scale, and constrain the learning direction of the wavelet kernel with self-supervised loss to achieve better image deblurring. Through extensive experiments on multiple real and synthetic datasets, we demonstrate that it outperforms existing state-of-the-art methods in terms of quality and efficiency of image restoration, especially with optimal deblurring performance and generalization in real scenes.

{
    \small
    \bibliographystyle{ieeenat_fullname}
    \bibliography{main}
}


\end{document}

%% file: preamble.tex
\usepackage[dvipsnames]{xcolor}
